\documentclass{article}

\usepackage{arxiv}

\usepackage[utf8]{inputenc} 
\usepackage[T1]{fontenc}    
\usepackage{hyperref}       
\usepackage{url}            
\usepackage{booktabs}       
\usepackage{amsfonts}       
\usepackage{nicefrac}       
\usepackage{microtype}      
\usepackage{lipsum}		
\usepackage{graphicx}
\usepackage{natbib}
\usepackage{doi}

\usepackage{amsmath}
\usepackage{natbib}
\usepackage{graphicx}
\usepackage{hyperref}
\usepackage{graphicx}
\usepackage{float}
\usepackage{caption}
\usepackage{subcaption}
\usepackage{microtype}
\usepackage{url}
\usepackage{booktabs}
\usepackage{xcolor}
\definecolor{darkblue}{rgb}{0, 0, 0.5}
\hypersetup{colorlinks=true, citecolor=darkblue, linkcolor=darkblue, urlcolor=darkblue}

\title{LayerNorm: A key component in parameter-efficient fine-tuning}

\author{ \href{https://orcid.org/0000-0001-6338-8469}{\includegraphics[scale=0.06]{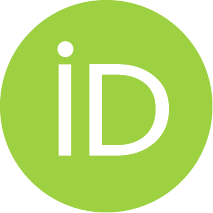}} \hspace{1mm}Taha ValizadehAslani\thanks{Corresponding author. E-mail address: taha.valizadehaslani@drexel.edu} \\
	Department of Electrical and Computer Engineering \\
	Drexel University \\
	Philadelphia, PA, USA \\
	\And
	\href{https://orcid.org/0000-0002-3805-1837}{\includegraphics[scale=0.06]{orcid.pdf}} \hspace{1mm}Hualou Liang  \\
	School of Biomedical Engineering \\
    Drexel University \\ 
    Philadelphia, PA, USA
}

\renewcommand{\shorttitle}{LayerNorm: A key component in parameter-efficient fine-tuning}

\hypersetup{
pdftitle={A template for the arxiv style},
pdfsubject={q-bio.NC, q-bio.QM},
pdfauthor={David S.~Hippocampus, Elias D.~Striatum},
pdfkeywords={First keyword, Second keyword, More},
}

\begin{document}
\maketitle

\begin{abstract}
	Fine-tuning a pre-trained model, such as Bidirectional Encoder Representations from Transformers (BERT), has been proven to be an effective method for solving many natural language processing (NLP) tasks. However, due to the large number of parameters in many state-of-the-art NLP models, including BERT, the process of fine-tuning is computationally expensive. One attractive solution to this issue is parameter-efficient fine-tuning, which involves modifying only a minimal segment of the model while keeping the remainder unchanged. Yet, it remains unclear which segment of the BERT model is crucial for fine-tuning. In this paper, we first analyze different components in the BERT model to pinpoint which one undergoes the most significant changes after fine-tuning. We find that output LayerNorm changes more than any other components when fine-tuned for different General Language Understanding Evaluation (GLUE) tasks. Then we show that only fine-tuning the LayerNorm can reach comparable, or in some cases better, performance to full fine-tuning and other parameter-efficient fine-tuning methods. Moreover, we use Fisher information to determine the most critical subset of LayerNorm and demonstrate that many NLP tasks in the GLUE benchmark can be solved by fine-tuning only a small portion of LayerNorm with negligible performance degradation.
\end{abstract}

\keywords{Parameter-Efficient Fine-Tuning \and LayerNorm \and Large Language Model \and Fisher Information}

\section{Introduction}

Transformer-based \citep{attention} Large Language Models (LLMs), such as Bidirectional Encoder Representations from Transformers (BERT) \citep{BERT}, Robustly optimized BERT approach (RoBERTa) \citep{RoBERTa}, and XLNet \citep{XLNet}, yield splendid performance for many natural language processing (NLP) tasks, outperforming traditional word embedding models, such as Word2Vec \citep{Word2Vec} and GloVe \citep{GloVe}. Such models are first pre-trained on a huge corpus of unlabelled text, and then are fine-tuned for a specific downstream task.

Despite their excellent performance, these models are computationally expensive for fine-tuning \citep{ExtremelySmallBERT,DiffPruning,BitFit,How-fine-fine-tuning,CompressingBERT} due to their large number of parameters. This cost grows with increasing the number of tasks learned \citep{How-fine-fine-tuning}. Moreover, such models with the large number of parameters in conjunction with limited labeled data for the downstream task are prone to overfitting and hence poor generalization performance for out-of-distribution data \citep{aghajanyan2021better,mahabadi2021variational,Child-Tuning}. One popular approach to this issue is to only train a small portion of the model, rather than performing a full fine-tuning. For instance, \cite{How-fine-fine-tuning} only trained 60 \% of BERT parameters. Recently, \cite{BitFit} only trained bias parameters of BERT, which reached results that are comparable with full fine-tuning. We hypothesize, however, that the bias may not necessarily be the optimal component of BERT for parameter-efficient fine-tuning, and similar/better performance could be obtained by training a smaller number of parameters if the optimal component is chosen.

In this paper, we use BERT as an example to test our hypothesis. First, we examine how different components of BERT change during the full fine-tuning and discover that LayerNorm is a key component in fine-tuning. Second, we show that LayerNorm possesses the maximum Fisher information among all the components of BERT. Third, we demonstrate that just training LayerNorm can reach the similar performance as only training bias, yet with one-fifth number of parameters. Finally, we show that a comparable performance can be obtained even with only a portion of the LayerNorm, where such a portion can be obtained from the information available in the down-stream task at hand, or other down-stream tasks. 

The rest of this paper is organized as follows. In Section \ref{search}, we demonstrate that LayerNorm is a key component in fine-tuning BERT. In Section \ref{method}, we present our method by only training LayerNorm and the results. Section \ref{discussion} is dedicated to the discussions. Related works are reviewed in Section \ref{related}. Finally, the conclusions and the future works are provided in Section \ref{Conclusion}. A detailed description of LayerNorm is provided in Appendix \ref{sec:LN}.

\section{A key component of BERT}
\label{search}

BERT consists of multiple layers, and in each layer there are different components, such as self-attention, feed-forward network, and LayerNorm. Our goal in this section is to pinpoint the most important component for fine-tuning.
\cite{How-fine-fine-tuning} demonstrated that in the BERT model, $L_1$ distance (See Appendix \ref{sec:distances} for definitions) between the pre-trained model and the fine-tuned model is significantly lower than the $L_1$ distance between two independent random initialed models, or the distance between parameters before and after pre-training. This indicates that during the process of fine-tuning, the model parameters only undergo small changes. Additionally, \textit{good} fine-tuned models exist that can have a small $L_0$ distance from the pre-trained model \citep{BitFit,How-fine-fine-tuning}. A small $L_0$ means that many parameters do not need to change. As such, we can achieve a good fine-tuning performance without training all the components. Therefore, the question is which components ought to be fine-tuned and which components do not need to change.

\subsection{Data}
We used the General Language Understanding Evaluation (GLUE) \citep{GLUE} dataset, which has been used in different studies \citep{Houlsby2019,DiffPruning,BitFit,Child-Tuning} as the standard benchmark in this field.
GLUE dataset consists of different tasks, namely, the Corpus of Linguistic Acceptability (CoLA) \citep{CoLA}, the Stanford Sentiment Treebank (SST2) \citep{SST}, the Microsoft Research Paraphrase Corpus (MRPC) \citep{MRPC}, the Semantic Textual Similarity Benchmark (STS-B) \citep{STSB}, the Quora Question Pairs (QQP) \citep{QQP}, the Multi-Genre Natural Language Inference Corpus (MNLI) \citep{MNLI}, the Stanford Question Answering Dataset (QNLI) \citep{QNLI}, and the Recognizing Textual Entailment (RTE) \citep{RTE1,RTE2,RTE3,RTE5}. We excluded the Winograd Schema Challenge (WNLI) \citep{WNLI} since the results for WNLI are unreliable \citep{Prasanna2020}. Many other studies have also excluded this task \citep{devlin-etal-2019-bert,BitFit,Child-Tuning}. Table \ref{tab:GLUE-metrics} in Appendix \ref{sec:GLUE-metrics} shows the metric employed for the evaluation of each task.

\subsection{Pipeline}
Low training cost can be achieved by only training a small subset of the model, which is equivalent to having a small $L_0$ distance between the pre-trained and the fine-tuned model.
Our goal is to find out which components in BERT must be frozen and which components must be trained in order to have a good performance and a low training cost.
To achieve the goal, we directly addressed the following tangigle question: During the process of full fine-tuning, which components of the model undergo significant changes? To proceed, we fine-tuned BERT-large-cased for different tasks in GLUE \citep{GLUE}. After fine-tuning, for each component, we compared the original value and the fine-tuned value of the parameters. For an array of all parameters in a component before fine-tuning (${C_{pre}}_i$) and after fine-tuning (${C_{fine}}_i$), where $i$ represents the index of different values in the component, and the size of the component is $n$, the change after fine-tuning, $D$, is defined as:

\begin{equation}
D= \frac{1}{n} \sum_{i=1}^{n} |{C_{fine}}_i-{C_{pre}}_i|
\end{equation}

which is equal to the $L_1$ distance normalized by the length of the vector, to compensate for different sizes in different components.

\begin{figure}[H]
     \begin{subfigure}[b]{0.5\textwidth}
         \includegraphics[width=\textwidth]{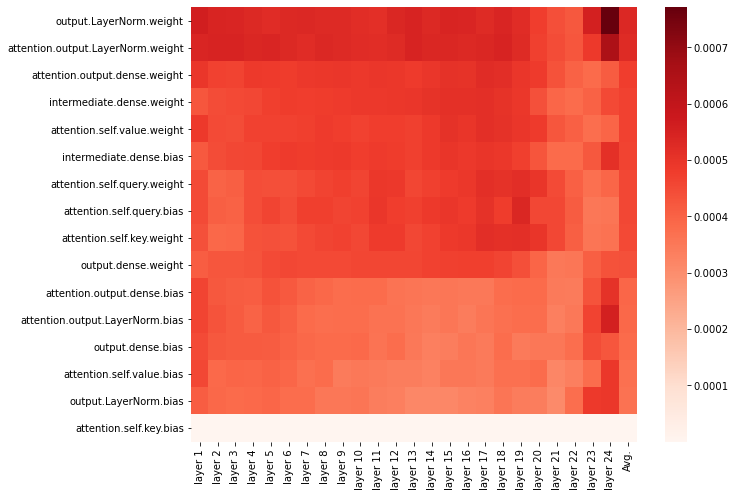}
         \caption{RTE}
     \end{subfigure}
     \begin{subfigure}[b]{0.5\textwidth}
         \includegraphics[width=\textwidth]{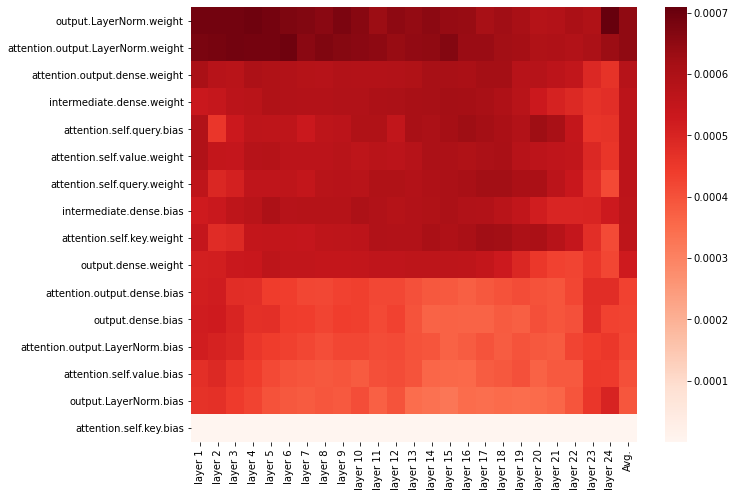}
         \caption{MRPC}
     \end{subfigure}
     \hfill
     
     \begin{subfigure}[b]{0.5\textwidth}
         \includegraphics[width=\textwidth]{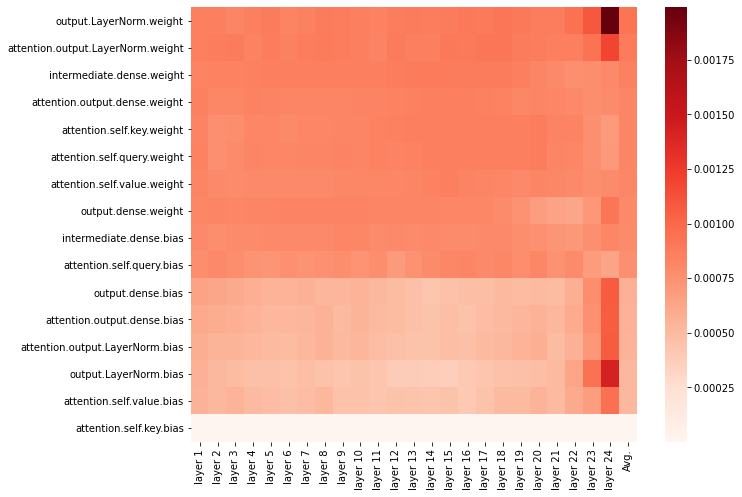}
         \caption{STSB}
     \end{subfigure}
     \begin{subfigure}[b]{0.5\textwidth}
         \includegraphics[width=\textwidth]{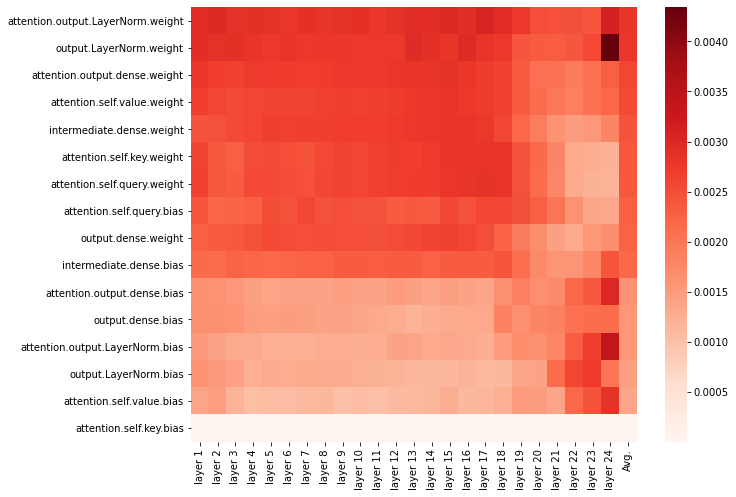}
         \caption{SST2}
     \end{subfigure}
     \hfill
    \caption{Heat map of change in each component after fine-tuning for different GLUE tasks.}
    \label{fig:HeatMapGLue}
\end{figure}

For all the components at different layers, we calculated $D$ and plotted the heat map for different GLUE tasks. These heat maps are presented in Figure \ref{fig:HeatMapGLue}. For most GLUE tasks, we observed that the most significant change happens in the output LayerNorm, which we simply call LayerNorm. 

\subsection{Effect of disabling LayerNorm}
Various studies have shown that Transformer-based models are, in general, remarkably robust to pruning \citep{16Heads1,CompressingBERT,Prasanna2020,chen2020lottery}, which means that removing parts of the model does not have a severe effect on its performance. Contrary to this, it has been shown by \cite{BERTBusters} that the performance of models from the BERT family degrades significantly if one component, LayerNorm, is disabled in the model. As an example, removing only 24 parameters in LayerNorm of RoBERTa \citep{RoBERTa} increases the loss of validation of WikiText \citep{WikiText} by nearly a factor of 4. 

\subsection{Search for the most important component using Fisher information}
Similar to \cite{Child-Tuning}, we used the Fisher information to choose which component to fine-tune and which to freeze. Fisher information is essentially an estimation of how much information a variable carries about a parameter of a distribution \citep{FisherInfo} and has proven to be a good metric to measure how important a certain parameter in a neural network is \citep{OvercomingCatastrophicForgetting,Child-Tuning}. In a dataset with $N$ samples, where $\mathbf{X_j}$ represents the $j$-th input sample and $y_j$ represents the $j$-th output, and $\omega$ represents parameters, the Fisher information for the $i$-th parameter can be represented as:

\begin{equation}
F(\omega^{(i)}) = \frac{1}{N} \sum_{j=1}^{N} {({\frac{\partial log(p(y_j|\mathbf{x_j}; \omega^{(i)}))}{\partial \omega^{(i)}}})^2}
\end{equation}

In each task, before running the fine-tuning, we showed the data to the model and calculated the gradient of all the parameters in all the components. Then we calculated the Fisher information of each parameter and obtained the average Fisher information of each component in each task. For each task, we normalized the Fisher information of each component by dividing it by the sum of all information in that task. The rationale for the normalization was to avoid the information of a task, where the total information is small, being overshadowed by a task where the total information is big, and ensure that all of the tasks equally contribute to the final result. After calculating the total information, we sorted the components in descending order, based on their total information in all tasks. The results are presented in Table \ref{tab:Fisher-components}. Again, we can see that ``output.LayerNorm'', which we call LayerNorm, has the maximum Fisher information, and ``attention.output.LayerNorm'', which we will call attention LayerNorm, comes the second.

\begin{table}[H]
\centering
\caption{BERT components sorted by the sum of normalized Fisher information in GLUE tasks.}
\label{tab:Fisher-components}
\begin{tabular}{|c|c|}
\hline
\textbf{Rank} & \textbf{Component}         \\ \hline
1 & output.LayerNorm           \\ \hline
2 & attention.output.LayerNorm \\ \hline
3 & attention.output.dense     \\ \hline
4 & attention.self.value       \\ \hline
5 & output.dense               \\ \hline
6 & attention.self.query       \\ \hline
7 & intermediate.dense         \\ \hline
8 & attention.self.key         \\ \hline
\end{tabular}
\end{table}

\section{Proposed method: Only training LayerNorm}
\label{method}

Based on the previous analysis and the observation of \cite{BERTBusters}, we hypothesized that freezing most of the BERT and only training LayerNorm would result in performance comparable to full fine-tuning. We provided several experiments to test this hypothesis.

\subsection{Fine-tuning results}
In this section, we reported the details of fine-tuning BERT-large-cased \citep{BERT}.
For each GLUE task, we tested the fine-tuning of the full model,  bias only (BitFit) \citep{BitFit}, LayerNorm only (our proposed method), and fine-tuning the same number of parameters as LayerNorm that were randomly selected. The random parameter experiment was performed as a control to show that the good performance of LayerNorm can not be obtained by any random choice of parameters. In each experiment, we tried 4 different learning rates on the validation set and selected the best. For full fine-tuning, we used the learning rates of $1*10^{-5}$, $2*10^{-5}$, $3*10^{-5}$, and $5*10^{-5}$, and for parameter-efficient fine-turnings (LayerNorm, BitFit, and random) we used $1*10^{-4}$, $4*10^{-4}$, $7*10^{-4}$, and $1*10^{-3}$. In all cases, we tried 20 epochs to select the best number of epochs. The development set results were obtained on our servers and are presented in Table \ref{tab:GLUE-BERT-large-dev}. Since for some GLUE tasks the true labels of the test data are held privately, we obtained test set results by submitting our test results to the GLUE benchmark website \citep{GLUEwebsite}. Test results are shown in Table \ref{tab:GLUE-BERT-large-test}.
The full model has 333,581,314 parameters, the BitFit method has 274,434 parameters and LayerNorm has 51,202 parameters, which is less than one-fifth of the number of parameters in the BitFit method.

\begin{table}[H]
   
\centering
\caption{Development set results of fine-tuning BERT large cased for different GLUE tasks, using different methods.}
\label{tab:GLUE-BERT-large-dev}
\begin{tabular}{|c|c|c|c|c|c|c|c|c|c|c|}
\hline
                   & \textbf{\% of full} & \textbf{QNLI} & \textbf{SST2} & \textbf{MNLI-m} & \textbf{MNLI-mm} & \textbf{CoLA} & \textbf{MRPC} & \textbf{STSB} & \textbf{RTE} & \textbf{QQP} \\ \hline
\textbf{Full}      & 100\%   & 0.9143 & 0.9335 & 0.8559 & 0.8567 & 0.6554 & 0.9239 & 0.9091 & 0.7653 & 0.8769 \\ \hline
\textbf{BitFit}    & 0.082\% & 0.9145 & 0.9278 & 0.8399 & 0.8457 & 0.6364 & 0.9183 & 0.9043 & 0.7473 & 0.8476 \\ \hline
\textbf{LayerNorm} & 0.015\% & 0.9072 & 0.9312 & 0.8285 & 0.8348 & 0.6412 & 0.9130 & 0.9039 & 0.7401 & 0.8361 \\ \hline
\textbf{Random}    & 0.015\% & 0.8975 & 0.9220 & 0.8113 & 0.8105 & 0.5851 & 0.8493 & 0.8822 & 0.6065 & 0.8391 \\ \hline
\end{tabular}
\end{table}

\begin{table}[H]
\centering
\caption{Test set results of fine-tuning BERT large cased for different GLUE tasks, using different methods.}
\label{tab:GLUE-BERT-large-test}
\begin{tabular}{|c|c|c|c|c|c|c|c|c|c|c|}
\hline
\textbf{}          & \textbf{\% of full} & \textbf{QNLI} & \textbf{SST2} & \textbf{MNLI-m} & \textbf{MNLI-mm} & \textbf{COLA} & \textbf{MRPC} & \textbf{STSB}  & \textbf{RTE}   & \textbf{QQP}   \\ \hline
\textbf{Full}      & 100\%                       & 0.917         & 0.918         &  0.849          &  0.841           & 0.590         & 0.890         &      0.872     & 0.706          & 0.696          \\ \hline
\textbf{BitFIt}    & 0.082\%                   & 0.912         & 0.928         &  0.841          &  0.838           & 0.559         & 0.877         & 0.860          & 0.687          & 0.687          \\ \hline
\textbf{LayerNorm} & 0.015\%                   & 0.910         & 0.926         &  0.831          &  0.828           & 0.542         & 0.871         & 0.865          & 0.682          & 0.669          \\ \hline
\textbf{Random}    & 0.015\% & 0.894  & 0.924  & 0.811  & 0.813  & 0.520  & 0.817  & 0.804  & 0.562  & 0.666  \\ \hline
\end{tabular}
\end{table}

Our results show that fine-tuning only LayerNorm can reach almost the same performance as BitFit suggested by \citep{BitFit}, yet with one-fifth of the parameters.

To check if there is any statistically signigcant difference in the performance between LayerNorm and BitFit, we ran the  Kruskal and Wallis (K-W) test \citep{Kruskal1952} between their results. Specifically, we compared BitFit and LayerNorm in two vectors, each having the 9 values from 9 different metrics in GLUE tasks. For the validation set, the K-W P-value was 0.56599 and for the test set, it was 0.62703. To increase the statistical power, we combined the results of validation and test to create $1\times18$ vectors.  For the combination of both validation and test results, the P-value was 0.54773. These tests indicate that the difference between groups is not statistically significant.

\subsection{Using only a portion of LayerNorm}
\label{sec:task-spec-mask}
\begin{figure}[h!]
    \centering
    \begin{subfigure}[b]{0.7\textwidth}
        \includegraphics[width=\textwidth]{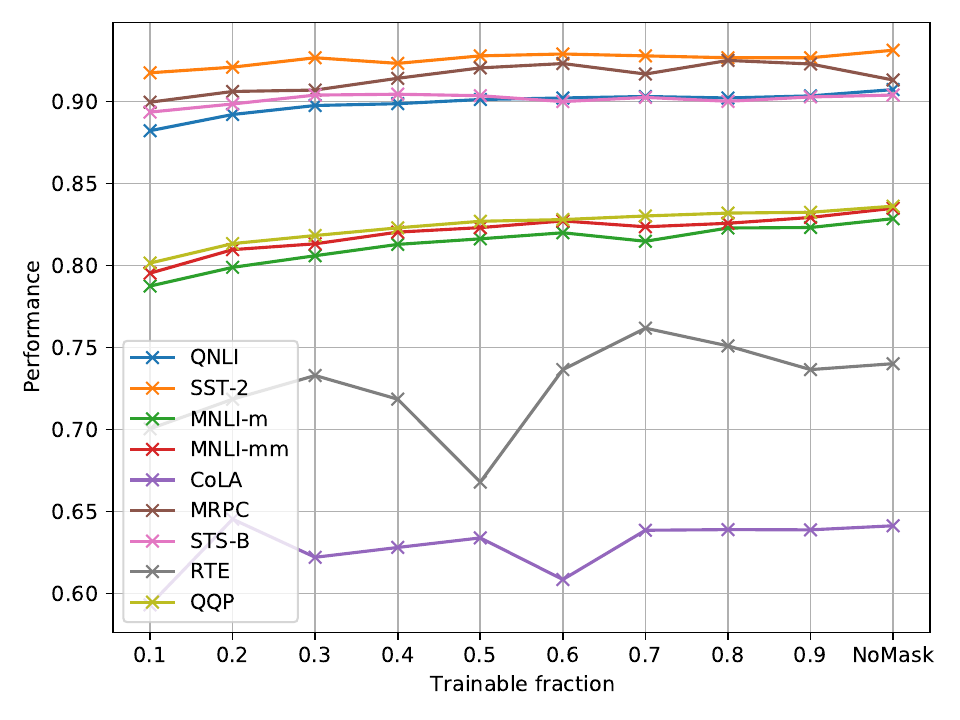}
    \end{subfigure}
    \caption{Validation results after training only a fraction of LayerNorm for different values of the trainable fraction.}
    \label{fig:Fisher-Mask}
\end{figure}

Next, we asked whether training all the parameters of LayerNorm are required. In other words, can we only train a subset of LayerNorm parameters and still maintain a good performance?
We used Fisher information to select a subset of LayerNorm parameters.
In each task, before running the fine-tuning, we calculated the gradient of all the parameters in LayerNorm and sorted the parameters based on their Fisher information. Then we only selected a fraction of parameters ($f$), where $0<f<1$, and fine-tuned the model based only on these parameters. For example, when $f=0.2$, only 20\% of LayerNorm parameters are fine-tuned and the remaining parameters are frozen. Similar to previous experiments, all other parameters in the components other than LayerNorm are frozen. The results of different tasks after freezing a portion of LayerNorm are presented in Figure \ref{fig:Fisher-Mask}.
The results show that only fine-tuning a small portion of LayerNorm parameters in some cases, such as QNLI, SST2, and STS-B slightly decreases the performance, but in other cases, such as MRPC and RTE, the performance is even improved. 

\subsection{Visualizing Fisher information of LayerNorm parameters}
\begin{figure}[h!]
    \centering
    \begin{subfigure}[b]{\textwidth}
        \includegraphics[width=\textwidth]{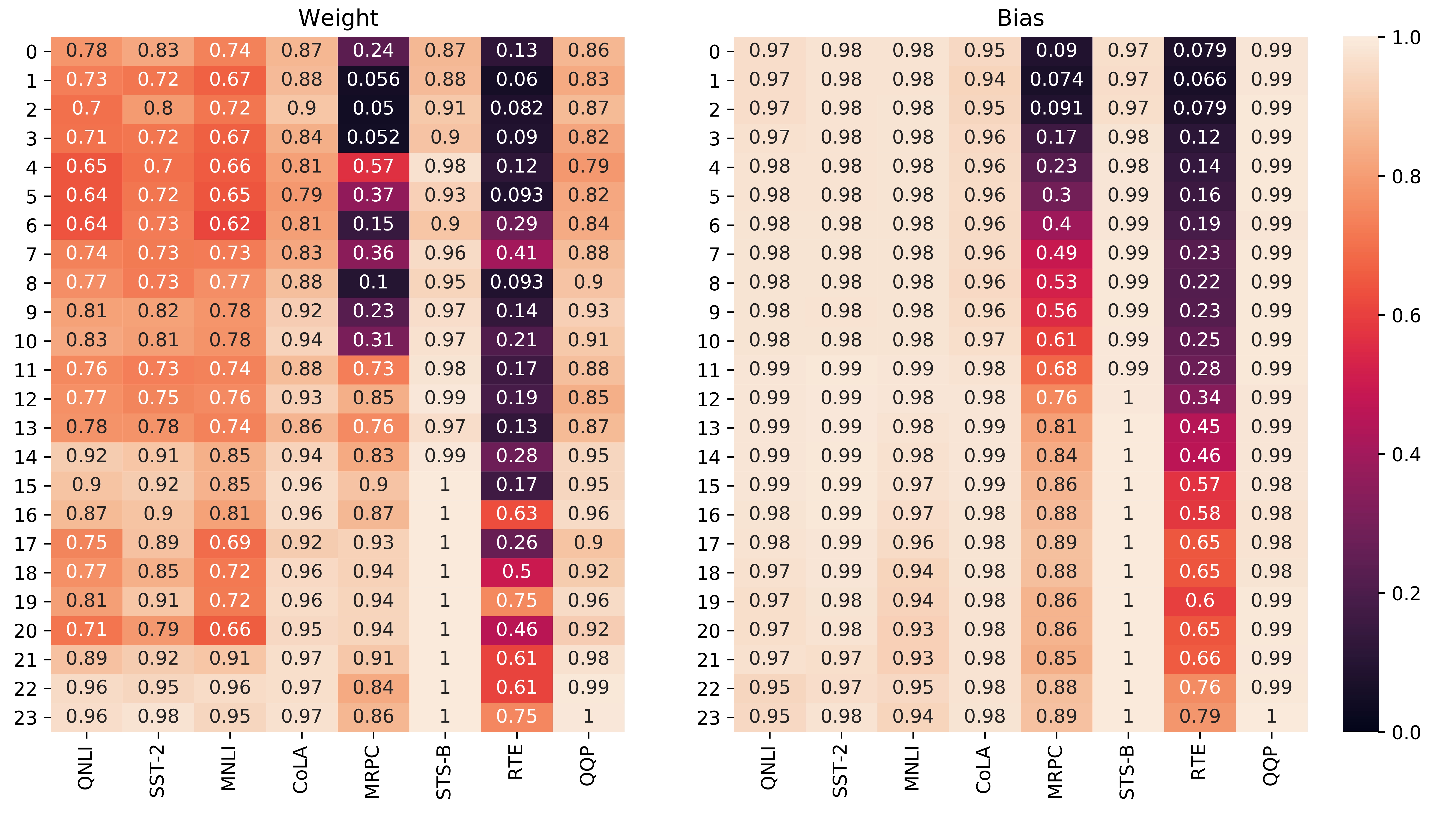}
    \end{subfigure}
    \caption{Heat map of Fisher information of LayerNorm in different layers. The X-axis is the task, the Y-axis is the Layer number.}
    \label{fig:LN-heatMap}
\end{figure}

In this section, we visualize Fisher information of different parameters of LayerNorm in various tasks.
We calculated the heat map of Fisher information of LayerNorm in different layers by summing the total Fisher information of each component. These heat maps are presented in Figure \ref{fig:LN-heatMap}. Since LayerNorm has the weight and bias sub-components, we plotted them separately. The task is shown in the X-axis, and the Layer number is shown in Y-axis.

Two findings can be observed from Figure \ref{fig:LN-heatMap}: (1) the LayerNorm of BERT contains more information in the final layers than the initial layers, and (2) there is more information in the bias terms than the weights.



\subsection{Global subset of LayerNorm}
\label{sec:GlobalSubsetofLayerNorm}
In section \ref{sec:task-spec-mask}, we fine-tuned only a portion of LayerNorm for each task in the GLUE benchmark. However, for each task, we used a different subset of the network. In other words, the selected subset was task-specific. As an alternative, in this section, we used a single subset of the LayerNorm parameters for all the tasks to make the selected subset task-independent.

\begin{figure}[h!]
    
        \centering
        \includegraphics[width=0.4\textwidth]{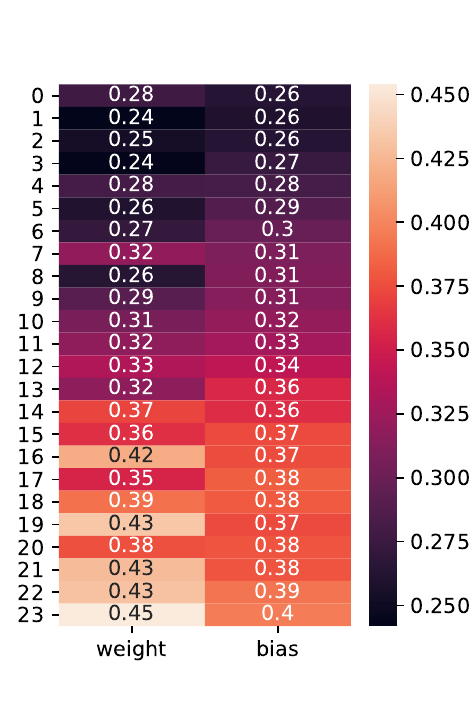}
    \caption{Heat map of Fisher information of LayerNorm in different layers. X-axis is the task and Y-axis is the Layer number. Left: Weight, right: Bias}
    \label{fig:LN-heatMap-Global}
\end{figure}

To find the global subset of the LayerNorm component, we calculated the Fisher information of each task separately, as described in section \ref{sec:task-spec-mask}. Then, eight Fisher information, corresponding to eight tasks, were normalized by dividing the information of each task by its sum. This step was performed, because otherwise in one task, the total amount of information could be higher than other tasks, which would make the information of one task overshadowed by another task. After normalization, the Fisher information of all tasks were added to create a single global information matrix. The heat map of this information is presented in Figure \ref{fig:LN-heatMap-Global}. This global information was used to create masks with different densities as described in section \ref{sec:task-spec-mask}. Validation results of training the model with the global subset of the LayerNorm are presented in Figure \ref{fig:Fisher-Mask-Union}, as labeled as global. For the sake of comparison, for each task, we also plotted the results of running the algorithm using the individual (task-specific) masks, labeled as individual. 

\subsection{Cross-validating the subset of LayerNorm}
\label{sec:GlobalSubsetofLayerNorm-CV}
\begin{figure}[h!]
    \centering
    \begin{subfigure}[b]{0.65\textwidth}
        \includegraphics[width=\textwidth]{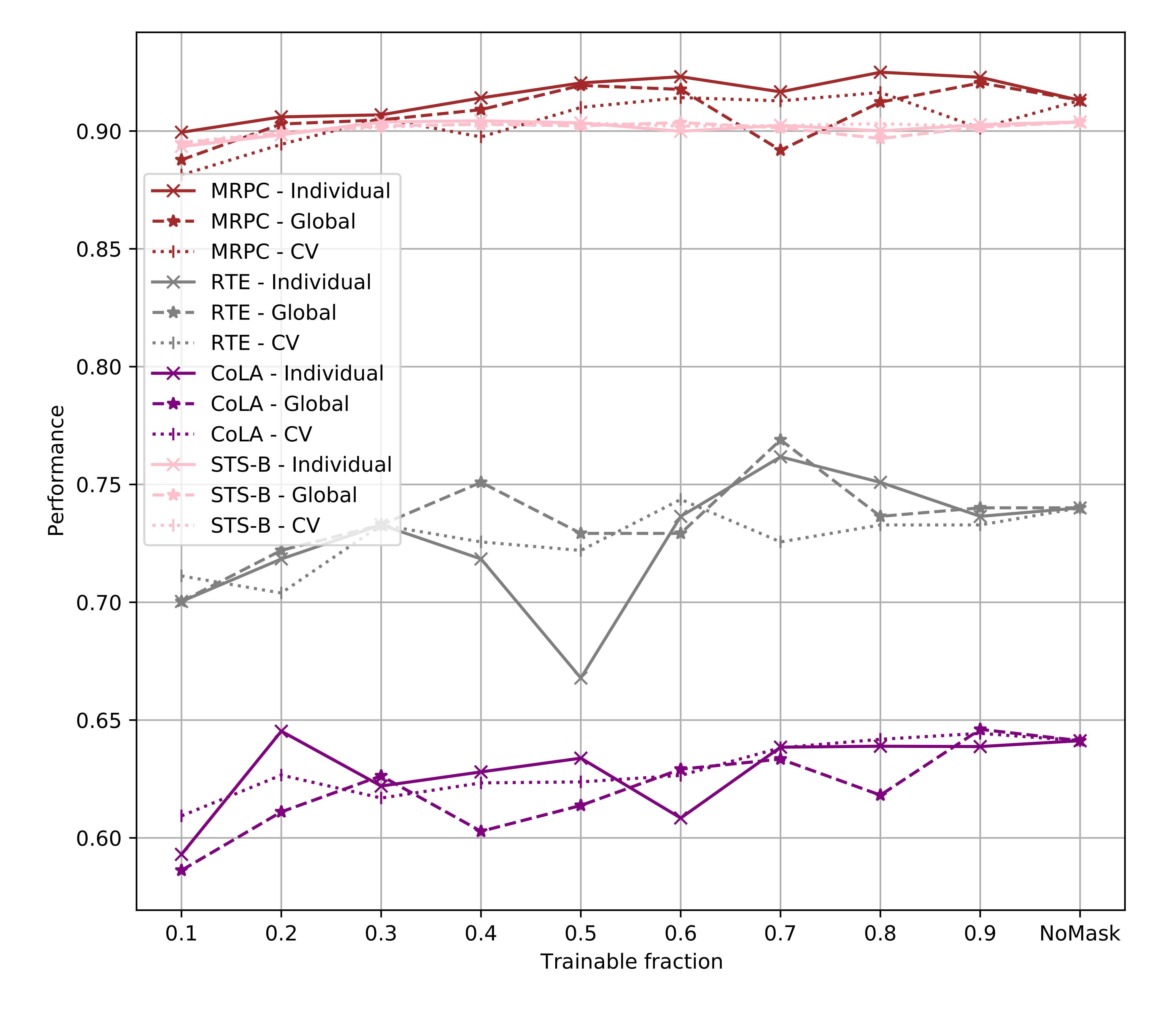}
    \end{subfigure}
    \caption{Validation results after training only a fixed fraction of LayerNorm for different values of the trainable fraction. For each task, the results of 3 experiments are plotted. Individual: The mask of each task is calculated based on information of that task. Global: The mask of each task is calculated based on information of all tasks. CV: The mask of each task has been calculated based on information of all tasks, excluding itself.}
    \label{fig:Fisher-Mask-Union}
\end{figure}

In section \ref{sec:GlobalSubsetofLayerNorm}, for each task, Fisher information of all eight tasks were used. To show that the global mask obtained is generalizable, we performed a cross-validation (CV) experiment: For each task, we used the Fisher information of all 7 other tasks to find the important subset of LayerNorm, excluding the target task itself. For example, the mask for MRPC was obtained by all tasks in GLUE except MRPC itself. In Figure \ref{fig:Fisher-Mask-Union}, the results of these masks are labeled as CV. The good performance of CV indicates that the mask obtained from other tasks can be generalized to a new task.

These results demonstrate that training a small subset of LayerNorm can achieve performance comparable to training the entire model. This is true even if the chosen subset is not task-specific, as observed in the global experiment, or when the subset is chosen without using the information of the task, as in the CV experiment.

\section{Discussion}
\label{discussion}

Although the Transformer-based models are robust against pruning \citep{16Heads1,CompressingBERT,Prasanna2020,chen2020lottery},
\cite{BERTBusters} have demonstrated that the performance from the BERT family degrades significantly if LayerNorm is disabled in the model. This is consistent with our findings: Not only does LayerNorm undergo greater change than other components during the fine-tuning, but also training LayerNorm alone or a small portion of it has comparable performance to the fine-tuning of the full model or other components with a much larger size of parameters.

Each parameter in LayerNorm is multiplied by a weight and then a bias term is added to it. This indicates that the bias andhas the same size as the weight for this component. In Figure \ref{fig:LN-heatMap}, we can see that for each task, bias terms have more information compared to the weight. This is consistent with the findings of \citep{BitFit}, where training only bias terms of BERT was proven to be effective.

In Figures \ref{fig:LN-heatMap} and \ref{fig:LN-heatMap-Global} we observe that there is more information in the LayerNorm of the final layers of BERT and less information in the LayerNorm of the initial layers. This trend more or less exists in all the GLUE tasks. The higher Fisher information is, the larger gradient. As such, during the fine-tuning the final layers have larger gradient and  will likely have more changes. This phenomenon has been observed in other studies \citep{WhatHappensToBERT,Yiwen_ADME,PharmBERT}.

\section{Related work}
\label{related}

In this section, we provide a brief overview of recent work related to parameter-efficient fine-tuning. In general, techniques proposed for parameter-efficient fine-tuning can be categorized into 5 groups: adding adaptors, adding prompts, model pruning, partial training, and low-rank decomposition. These groups are explained  as follows.

\textbf{Adding adaptors} is to add trainable modules, called adaptors, into the original frozen model and only training the adaptors. Examples of such categories are \cite{Adapters, DiffPruning, Mahabadi2021ParameterefficientMF}. \cite{Adapters} suggested injecting adapters between layers of the pre-trained network. \cite{DiffPruning} proposed adding a sparse, task-specific difference-vector (diff-vector) to the pre-trained network.

\textbf{Adding prompts} involves prepended new tokens to the input text and only training the embeddings of these prompt tokens \citep{lester-etal-2021-power,razdaibiedina2023residual}. In such methods, the backpropagation is forced to apply the changes to the vectors corresponding to the soft prompt because the core model is frozen.

\textbf{Model pruning} is to remove certain weights from the network \citep{braindamage,brainsurgeon,Han2015LearningBW,Xia2022StructuredPL,xia2023sheared,sun2024a}. To decide which weights should be pruned, \cite{Han2015LearningBW} removed weights with low magnitude while \cite{sun2024a} evaluated each weight by the product of its magnitude and the norm of the corresponding input activation.

\textbf{Partial training} is only training a subset of the model \citep{Lee2019WhatWE,BitFit,Child-Tuning}. \cite{Lee2019WhatWE} only trained one-fourth of the final layers of BERT and RoBERTa. \cite{BitFit} suggested only fine-tuning the bias parameters. \cite{Child-Tuning} used Fisher information to select the most important parameters. Our proposed method falls within this category.

\textbf{Low-rank decomposition} methods utilize low-rank decomposition to approximate model update during fine-tuning \citep{hu2022lora, zhang2023adalora, woo2022fedpara, kopiczko2024elora, liu2024dora}. \cite{hu2022lora} proposed Low-Rank Adaptation (LoRA). LoRA approximates the weight change of fine-tuning as the product of two low-rank matrices, where rather than training the whole model, only these low-rank matrices need to be trained. \cite{liu2024dora} decomposed the pre-trained weight into two components, magnitude and direction, and used LoRA for the directional adaptation. These methods are similar to partial training because they apply a small modification to the weights of the pre-trained model.

\section{Conclusions and future work}
\label{Conclusion}

In this paper, we first examined all the components of BERT when fine-tuned for different GLUE tasks and showed that LayerNorm undergoes more changes after fine-tuning compared to other components. This is consistent with the findings of \cite{BERTBusters}, where it was shown that, unlike other components, disabling LayerNorm has a dramatic negative effect on the performance of BERT. We then showed that only fine-tuning LayerNorm has a comparable performance to  Bitfit, proposed by \cite{BitFit}, in spite of being more sparse. Finally, using Fisher Information, we were able to select the important subsets of LayerNorm parameters and demonstrated that with slightly performance degradation, comparable results can be obtained by only fine-tuning as low as only 10\% of LayerNorm parameters, which is one hundred thousandths of the BERT model. 

In our analysis, we focused on the layer normalization, which is the popular method for normalization in the realm of NLP. However, in other fields, such as computer vision, batch normalization (see Appendix \ref{sec:LN}) has been widely adopted \citep{158}. Applying the parameter-efficient training to batch-normalization can be employed as an extension of our work, and hence can make the training of batch-normalization models more computationally efficient.

\bibliographystyle{colm2024_conference}


\appendix
\section{Normalization in neural networks}
\label{sec:LN}


Generally, during the training of a deep neural network, adjustments in the parameters of a certain intermediate layer will cause a change in the distribution of the input to the next layer. This, phenomenon, called internal covariate shift, slows down the process of training by requiring a lower learning rate and a carful parameter initialization \citep{batch_norm}.

\subsection{Batch normalization}

One solution to this problem is batch normalization, which computes the mean and variance of the inputs to a neuron across a mini-batch of training examples. These statistics are then used to normalize the inputs to that neuron for each training sample. This simple modification makes the model robust against internal covariate shift and significantly reduces the training time \citep{batch_norm}.

\subsubsection{Shortcomings of batch normalization}

In spite of its effectiveness, batch normalization has multiple shortcomings:

\begin{enumerate}

    \item \textbf{Dependency on mini-batch size}: Batch normalization relies on the mean and variance of the inputs across the mini-batch. This dependency can introduce significant variability in the normalization process, especially with small mini-batch sizes, which can be problematic for tasks that require small batches due to memory constraints \citep{157}.
    \item \textbf{Performance in recurrent neural networks (RNNs)}: Batch normalization is less effective in RNNs due to the sequential nature of the data and the complications arising from applying normalization across time steps \citep{157}.
    \item \textbf{Inference complications}: Batch normalization requires maintaining running averages of the mean and variance during training to use for normalization during inference. This can complicate the model's deployment, especially in models that see a significant shift in the input distribution at inference time. Unlike batch normalization, layer normalization carries out the same process at the training and inference phase \citep{157}.
    \item \textbf{Batch Dependency}: Since batch normalization normalizes inputs based on the batch's statistics, it introduces a form of dependency between training examples within the batch. This can affect the model's ability to generalize, especially for tasks where the independence of examples is crucial. This problem becomes particularly severe in NLP problems, where statistics of different batches are significantly different \citep{158}.
    \item \textbf{Inability to maintain criticality}: In \citep{159}, it was shown that batch normalization can cause gradient explosion. Generally, in a neural network, two undesired situations are exponentially growing co-variance and exponentially decaying covariance. The critical point between these two undesired situations is the point where the network has a perfect self-similarity of the co-variance and preserves it through the training from layer to layer \citep{160}. This desired situation is called criticality \citep{160}. Similar to Dropout \citep{161}, batch normalization destroys criticality.

\end{enumerate}

\subsection{Layer Normalization (LayerNorm)}

\subsubsection{Description}

\begin{figure}[h!]
    \centering
    \begin{subfigure}[b]{0.9\textwidth}
        \includegraphics[width=\textwidth]{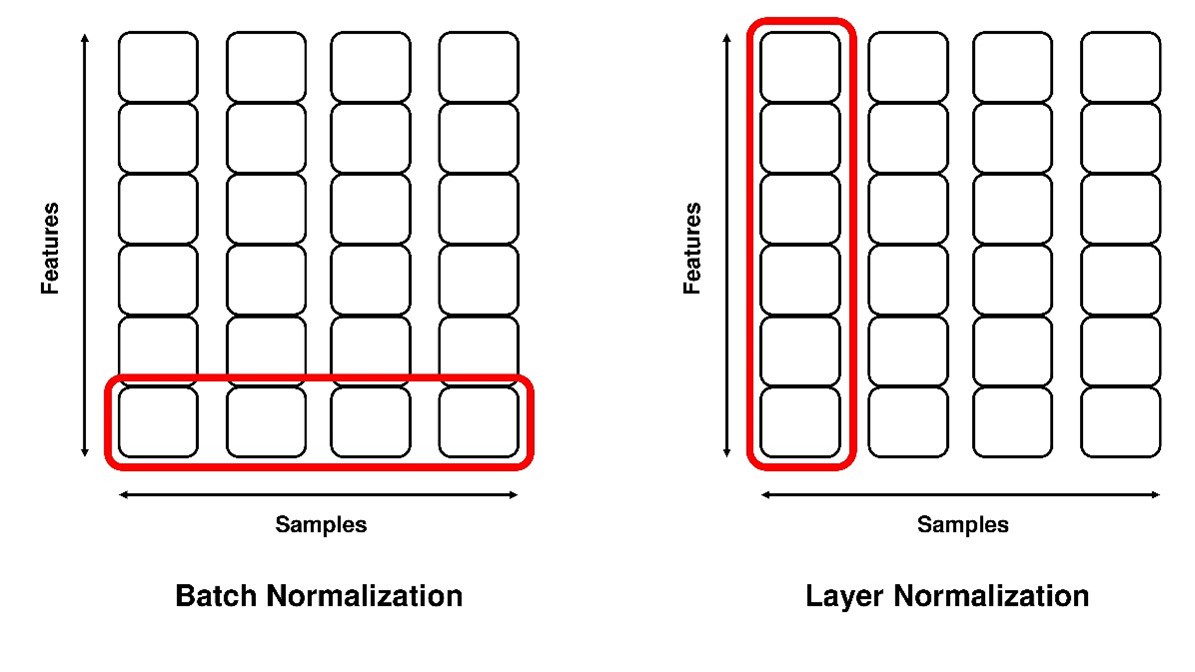}
    \end{subfigure}
    \caption{Unlike batch normalization, in layer normalization, normalization is performed across all features in the layer, not the same features in different samples of the batch..}
    \label{fig:BN_LN}
\end{figure}

Unlike, batch normalization, in LayerNorm, normalization is performed across the layer, not the batch, and all the hidden units in the same layer share the same normalization parameters for mean and variance \citep{162} (see Figure \ref{fig:BN_LN}). As a result, normalization does not depend on the batch size and can be done with any batch size (including 1) \citep{157}.

LayerNorm works well in RNNs \citep{157}. Additionally, the process of layer normalization is exactly the same in training and inference \citep{157}. Unlike batch normalization, which destroys criticality, LayerNorm, maintains criticality because proper stacking of LayerNorm leads to an architecture that is critical for any initialization \citep{163}.
It was demonstrated by \cite{164} that LayerNorm plays a crucial role in controlling the gradient scales and the effect of the location of LayerNorm on the gradient was investigated. In \citep{165}, the problem of uneven gradient magnitude mismatch in different layers was mitigated by adding 3 normalization operators to each layer. Alternative versions of LayerNorm have also been proposed to improve gradient propagation \citep{158}, \citep{166}, \citep{167}. \cite{166} suggested removing bias and gain parameters of LayerNorm to prevent over-fitting. \cite{167} proposed variance-only LayerNorm, where normalization is done without subtracting the mean in equation (4). LayerNorm has also been employed in RNNs \citep{168}.
Next, a more technical description of LayerNorm is provided.

\subsubsection{Technical details}

LayerNorm normalizes the outputs of both self-attention and linear layers \citep{LayerNormalization}. For an input of the $i$-th layer, $\mathbf{x_i}$, of size $H$, where each element is represented by $x_{ij}$, LayerNorm first computes mean $\mu_i$ and variance ${\sigma_i}^2$ across the features:

\begin{equation}
\mu_i=\frac{1}{H} \sum_{j=1}^{H} {x_{ij}}, \; \; {\sigma_i}^2=\frac{1}{H} \sum_{j=1}^{H} {({x_{ij}-\mu_i})^2}
\end{equation}

After calculating the above values, the inputs are normalized based on these values:

\begin{equation}
\label{eq:LN}
\hat{x}_{ij}=\frac{x_{ij}-\mu_i}{\sqrt{\sigma^2+\epsilon}}
\end{equation}

Note that $\epsilon$ is only used to avoid zero-division.
Then, the normalized value go through an affine transformation, which contains the learnable parameters $\boldsymbol\omega$ and bias $\mathbf{b}$:

\begin{equation}
\mathbf{y_i}= \boldsymbol\omega \odot \mathbf{\hat{x}_{i}} + \mathbf{b}
\end{equation}

where $\odot$ is Hadamard (element-wise) multiplication.

\section{Distance definitions}
\label{sec:distances}
For 2 vectors, like $V_1$, and $V_2$, of size $n$, $L_0$ distance is defined as the number of non-zero elements in ${V_{1}}_i-{V_{2}}_i$ where $i$ indicates element index. This is a similar idea to Hamming distance \citep{waggener1995pulse}.

$L_1$, or Manhattan, distance, between 2 such vectors is defined as:
\begin{equation}
L_1 = \sum_{i=1}^{n} |{V_{1}}_i-{V_{2}}_i|
\end{equation}

\section{Metrics for GLUE results}
\label{sec:GLUE-metrics}
Table \ref{tab:GLUE-metrics} shows the metric used for each task.

\begin{table}[H]
\centering
\caption{Metrics used to evaluate GLUE Benchmark.}
\label{tab:GLUE-metrics}
\begin{tabular}{|c|c|}
\hline
\textbf{Task}  & \textbf{Metric}             \\ \hline
QNLI  & accuracy                             \\ \hline
SST-2 & accuracy                             \\ \hline
MNLI  & Matched accuracy/mismatched accuracy \\ \hline
CoLA  & Matthews correlation                 \\ \hline
MRPC  & F1                                   \\ \hline
STS-B & Spearman correlation                 \\ \hline
RTE   & accuracy                             \\ \hline
QQP   & F1                                   \\ \hline
\end{tabular}
\end{table}



\end{document}